# Machine Learning Epidemic Predictions Using Agent-based Wireless Sensor Network Models

Chukwunonso Henry Nwokoye
ANTS Research Lab,
Ontario Tech University,
Oshawa, Canada
HenryChukwunonso.Nwokoye@ontariotechu.ca

Blessing Oluchi Iloka
University of Hertfordshire,
England,
United Kingdom
Boluchi23@gmail.com

Sharna Waldron
Department of Cybersecurity,
ABM College
Toronto Campus, Canada.
sharnawaldron@gmail.com

Peace Ezzeh
School of Science Education,
Federal College of Education
(technical) Asaba, Nigeria.
peace.ezzeh@fcetasaba-edu.ng

*Abstract*— **The lack of epidemiological data in wireless sensor networks (WSNs) is a fundamental difficulty in constructing robust models to forecast and mitigate threats like viruses and worms. Many studies have looked at different epidemic models for WSNs, focusing on the manner in which malware infections spread given the network's specific properties, including energy limits and node mobility. In this study, an agent-based realization of the susceptible-exposed-infected-recovered-vaccinated (SEIRV) mathematical model was employed for machine learning (ML) predictions. Using tools such as Netlogo's BehaviorSpace and Python, two epidemic synthetic datasets were generated and prepared for the application of several ML algorithms. Posed as a regression problem, the infected and recovered nodes were predicted, and the performance of these algorithms is compared using the error metrics of the train and the test sets. The predictions performed quite well, with low error metrics and high $R^2$ values (0.997, 1.000, 0.999, 1.000), indicating an effective fit to the training set. The validation values were lowered (0.992, 0.998, 0.971, and 0.999), as is typical when evaluating model performance on unknown data. Judging from the recorded performances, support vector, linear, Lasso, Ridge, and ElasticNet regression were among the worst performing algorithms, while Random Forest, XGBoost, Decision Trees, and K nearest neighbor had the best model performances.**

I. INTRODUCTION

In recent years, the globe as we know it has been changing due to breakthroughs in numerous linked innovations including smart electrical grids [1], the IoT, long-term evolution, 5G connectivity [2] and cyber physical systems [3] such as wireless sensor networks (WSN). Essentially, the Internet gives inherent assistance to the ever-changing cyberspace, with its continuous operation presenting a torrent of opportunities, convenience, and advantages. Nevertheless, these developments carry the risk of malicious and hostile attacks from black hat attackers who want to disrupt users' or organisations' regular use of the world wide web for meaningful purposes [4]. The rapid increase in malware spread poses significant threats to security risks owing to the interchange of large volumes of private data/information through devices with limited resources and across an insecure Internet utilising diverse technological platforms and protocols. In reality, such destructive programs with extremely detrimental dispositions threaten the cyberspace's existence and profitability. More so, the sources of attacks in communication networks include ransomware, worms, rootkits, viruses, trojan horses, and ransomware [5]. The proliferation of malware has proven destructive to organisations and enterprises, and increasingly lethal facts has been uncovered, indicating a dark shift in cybercrime towards attacking on educational institutions, municipal agencies, and various other severely short-staffed and overworked government agencies [6]. Our interest here is the propagation of these malware variants in WSNs," which are network of sensor nodes that collectively monitor and perhaps control the sensor field, thus facilitating interactions between individuals or computer systems as well as the field of interest" [5]. WSNs are recently used precision farming, battlefield sensing and underwater monitoring. Given the increasing prevalence of malware in sensor networks, malware identification is required to handle the fast-increasing pace and amount of newly developed malware [7]. Aside the use of anti-malware software, epidemiological models have additionally been utilised to describe the propagation behaviours of malware as well as combat the ongoing cyber-attack on information and communication technology [5]. Epidemic techniques have their roots in public health and epidemiology in which contagious effects on populations are assessed with the goal to comprehend spread patterns. Cybersecurity professionals have discovered significant parallels connecting pathogens that cause diseases in biological systems and virtual infections in communication/technological networks. Epidemic models of various kinds are mostly used due to the challenging nature of obtaining the real epidemiological data concerning sensor networks [8]. The aim of the study is to conduct machine learning (ML) predictions using data sets generated using agent-based models. The ML algorithms employed here include histogram gradient boosting (HGB), random forest (RF), decision trees (DT), support vector regression (SVR), linear regression (LiR), lasso regression (LaR), ridge regression (RiR), k nearest neighbour (KNN), elastic net regression (ENR). The paper is arranged as follows: related works are presented in Section 2, while the Section 3 contains the model. Section 4 contains the results of the study whereas Section 5 contains the conclusion and future directions.

II. RELATED WORKS

The brief review of related literature here include mathematical and agent-based models as well as literature on predictions done using ML and neural networks. The following are WSN mathematical models: Wu, et al. [9] proposed an susceptible-infected-susceptible (SIS) epidemic model considering sleep scheduling necessary for the efficient use of sensor energy. Liu, et al [10] used the Susceptible, Infected, Susceptible, Low-Energy (SISL) model to represent virus spread in WSNs. Liu, et al. [11] use the Susceptible, Infected, Low-energy, Susceptible model under pulse charging model to depict malware spread and containment. Note that these papers reviewed epidemic mathematical modes on WSN [5], [12]. The advantage of the agent-based approach is that it takes the mind away from stability analysis performed in equation-based models and concentrates on the behaviours of the compartments. We did not find several agent-based WSN models; however, we reviewed other types of networks. Kotenko [13] employed the agent-based approach to model DDoS attacks characterizing agents' team structures and action plans. Kotenko [14] employed both multi-agency and discrete event to model DDoS, botnets and worms and simulated network protocols at the packet level. The outbreak of malware was modeled using the agent approach for plans that are both organized and unorganized. Niazi & Hussain [15] used the agent-based method so as to characterize self organization in peer to peer (P2P) networks. Mojahedi & Azgomi [16] considered the network topology and time lag to model the time of infection for P2P worms. Contrarily, Wasti [17] did not consider epidemic theory because his model considered routing, medium access control, resource allocation, and cognitive radio. Like Mojahedi, Niazi & Hussain [18] applied boid animal agent model to represent the WSN, unfortunately, this model is not based on epidemic theory.

Shone, et al. [19] used the NSL-KDD and KDD Cup '99 datasets and the auto-encoder for learning features in order to evaluate detection accuracies for human interactions. Chawla [20] employed ML algorithms for evaluation of stability and traces of a real IoT network. Rhode, et al., [7] conducted behavior analyses and detection of malware (virus, worms, trojan, rootkit, etc.) using the following ML algorithms (RF, MLP, KNN, SVM, DT, AdaBoost, NaiveBayes, GradientBoostedDecisionTrees) and RNN. Like the above study, Xiaofeng, et al., [21] employed both ML and deep learning approaches for the detection of malware and the combination architecture with combined algorithms yielded an accuracy of 99.3%. Hijazi, et al. [22] employed deep learning methods in order to build an enhanced network capable of attack identification. Kang, et al. [23] used Word2vec- based LSTM method for the analyses of application programming interface functions and opcodes. Fang, et al. [24] predicted rates of cyber attacks the bi-directional version of the LSTMRNN, and it was found that several error metrics (MAPE, MSE, PMAD and RMSE) was better than autoregressive integrated moving average (ARIMA). Thamilarasu & Chawla [25] used ML algorithms for security issues in IoT networks, evaluating network traces and traceability. Ren, et al., [26] focused on the detection of anomalies in cases of data sampling, few records and selection of features using the UNSW-NB15 dataset. Boukhalfa, et al. [27] explore malware intrusions using LSTM in order to discover a two-fold method of defending against such intrusions. Almseidin, et al [28] explored the space for the metrics of false positives and negatives using ML classifiers and the decision tree classifier displayed the highest accuracy. Wuke, et al. [29] attempted to solve the issues of failures in extracting feature that are both abstract and representative using both the multi-layer extreme learning machine and support vector regression approaches. Kim [30] conducted performance evaluation for the rates of detection for attacks on WiFi networks using both selection and deep-feature extraction techniques. Wang, et al., [31] used a topology-aware communication epidemic model for different worms to analyse the dynamical behaviour of propagation is a competitive manner, considering algorithms such as RegularNN and SVM. Alomari et al. [32] employed dense and LSTM-based to address both high-dimensional data and malware detection systems with high performance in order to show that there exists degradation of performance between 0.07% to 5.84%, and 3.79% to 9.44% for the first and second dataset, respectively.

For LSTM, the high-performance accuracies derived from several intrusion detection studies is enough reason to guarantee use. For instance, a 99.968% detection accuracy was recorded in Bediako [33], while a 97.59% prediction accuracy was recorded in Kang, et al. [23] with 0.5% higher performance. In the study by Kang, et al. (2019) [23], "the prediction accuracy of the proposed method was initially 77.11% and grew to a final value of 97.59%". Boukhalfa, et al. (2020) recorded an accuracy reaching up to 99.98% and 99.93% in the LSTM study, justifying that it is more accurate compared to KNN, Trees and SVM. The LSTM-RNN-K proposed by Kim, et al. [34], with Softmax classifier achieved an accuracy of 96.930% and 98.110% respectively. The LSTM-RNNS proposed by Staudemeyer [35] with the LSTM-RNN classifier attained an accuracy of 93.8%. Aside from this study [31], none of the reviewed papers considered epidemic models for dataset generation and application of ML/DL algorithms, hence the need for the work herein.

### III. MODEL

The WSN mathematical and agent model used herein was initially developed by Nwokoye & Umeh [8]. In the model, worm attacks WSNs are characterized using the Susceptible-Exposed-Infectious-Recovered-Susceptible with Vaccination class (SEIRV), whose properties are listed below. In the Wireless Sensor Network, sensors are classified as in any of the above-mentioned compartments, represented by $S(t), E(t), I(t), R(t),$ and $V(t)$. Consequently, $S(t) + E(t) + I(t) + R(t) + V(t) = N(t)$. The parameters and their meanings are as follows: sigma is the density of distribution density, r is range of transmission, and the product of sigma, pi and r becomes the rate of contact with an infectious node for the effective transfer of malware infection. lambda is the rate at which sensors are added to the population, beta is the contact rate of infection, tau is the rate of death due consequent upon on failures (software/hardware), omega is the crashing rate resulting from malware (worm) attack, theta is the transmission from E to I compartment, nu is the recovery rate of sensors, phi is the rate at which R nodes transition to S compartment, rho is the transmission rate from V to S compartment, and xi is the transmission from the V to the S compartment. Finding the solutions for the existent equilibrium states have been addressed in the paper [8], however, it is not the focus of this study. The above-described mathematical model is presented below and its equivalent agent-based model is depicted as Figure 1 below.

$$\frac{dS}{dt} = \lambda - \beta SI\sigma\pi r_0^2 - \tau S - \rho S + \varphi R + \xi V$$
$$\frac{dE}{dt} = \beta SI\sigma\pi r_0^2 - (\tau + \theta)E$$
$$\frac{dI}{dt} = \theta E - (\tau + \omega + v)I \qquad (1)$$
$$\frac{dR}{dt} = vI - (\tau + \varphi)R$$
$$\frac{dV}{dt} = \rho S - (\tau + \xi)V$$

*A. Synthetic Dataset Generation with Implied Rates (First Model)*

We generated a total of 5 csv sheets using NetLogo's BehaviourSpace tool, corresponding to 5 different experiments, where the widgets were tweaked using several values. Although, the agent model contained several widgets, we focused on the following as well as their corresponding values used for the experiments: Infectiousness (20, 40, 60, 80, 100); Worm-duration (10, 30, 50, 70, 90); Duration-of-exposure (40, 45, 50, 55, 60); Number-of-nodes (200, 400,

600, 800, 1000); Chance-recover (100, 80, 60, 40, 20); and Chance-of-vaccination (90, 70, 50, 30, 10).

Fig. 1. First SEIRV Model

Our intention here is to understand the impact of infectiousness, worm-duration, duration-of-exposure, number-of-nodes, chance-recover and chance-of-vaccination on the susceptible, exposed, infected, recovered and vaccinated nodes. On the one hand, increasing infectiousness, worm-duration, and duration-of-exposure mimics a weak defense strategy by an organization. It may also imply the lack of an anti-malware scheme or the effect of an anti-malware software that is not updated with recent malware signatures, heuristic and behavioural changes. On the other hand, increasing chance-recover and chance-of-vaccination mimics a strong defence strategy against malware spread. Additionally, it may also be equivalent to constantly updating the malware software with new signatures. Note also that the max-pxcor, min-pxcor, max-pycor, minpycor values for all the experiments are -13, 13, -12 and 12, respectively. Figure 2 below shows an example of running experiment 1 using the BehaviourSpace tool. The behavioural changes of the model is depicted in the result section. Afterwards, we generated another dataset where the max-pxcor, min-pxcor, max-pycor, min-pycor values for all the experiments was changed to -35, 35, -35 and 35, respectively.

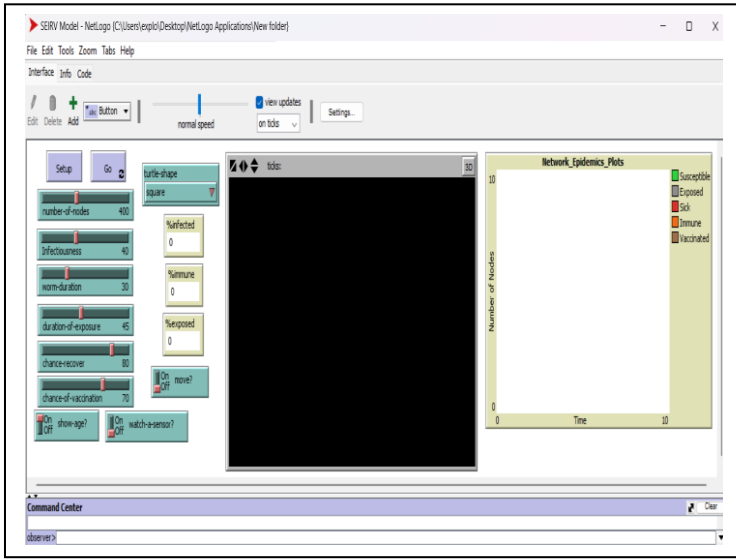

### B. Preprocessing and Training (First Model)

Firstly, we renamed some column names for uniformity, i.e., sick to infected, immune to recovered. Secondly, we merged Excel sheets of several experiments after installing and importing several important libraries. Then the ydata Profiling Package was installed, imported, and used to find issues with the dataset. Consequently, there were no missing values. However, the alerts generated from the ydata profiling report mentioned the exposed column as the column wherein other features are highly correlated with, so this column was dropped. This reduced the columns to an SIRV model. This evidently shows the problem of high correlation with synthetic datasets, and it can be traceable to the dataset generation approach, which is agent-orientated programming. Note that even after dropping the exposed column (feature), the report still showed high correlation for the remaining features. Furthermore, StandardScaler was used to adjust the features

Identify applicable funding agency here. If none, delete this text box.

with mean 0 and standard deviation 1, while the Synthetic Minority Over-sampling Technique (SMOTE) method was utilised to balance the target categories by creating artificial samples in the minority class, resulting in a more balanced dataset. SMOTE assists to balance an unequal dataset by creating synthetic examples for the minority class. This method guarantees all features are consistent and the dataset itself is balanced, which improves the performance of the model while handling unbalanced data. However, SMOTE failed, therefore, we used the weighted loss function (WLF) for balancing the dataset. With WLF, we ensure that the less frequent target variables are properly learnt, providing greater weights to them during model training.

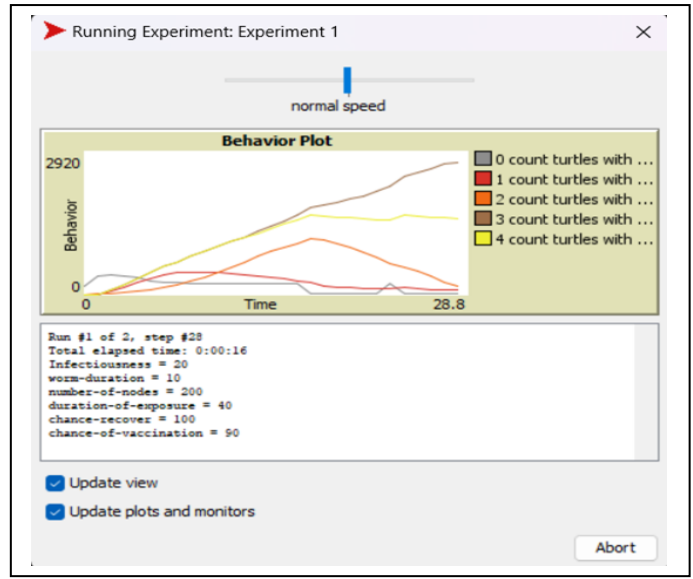

### C. Synthetic Dataset Generation for SEIRV Model with Explicitly coded Rates (Second Model)

We employed a different approach for the SEIRV epidemic dataset generation here and this is because it was observed from the profiling report that the Exposed, Infected, Recovered and Vaccinated columns are highly correlated. Our aim here is to generate a dataset that posses unique values for the existing compartments (susceptible, infected, exposed, recovered and vaccinated). Here, all the rates were expressed and the sliders can be used to select values of choice. The twelve rates proposed SEIRV model was explicitly coded alongside a slider functionality to enable selection of choice rates. This is unlike the NetLogo case above, where some of the rates are implied and not explicitly represented.

Fig. 2. BehaviourPlot for the First SEIRV Model

### D. Preprocessing and Training (Second Model)

Firstly, we installed and imported necessary packages and applied the ydata profiling on this dataset so as to generate alerts concerning areas of treatment. Secondly, we used the ydata profiling report to discover issues with the dataset. Consequently, the alerts showed that the existing compartments now have unique values and not highly correlated values. Thirdly, we dropped the rates as our focus is on the compartments and rounded off the two decimal places. Then, we set the target and feature variables of the dataset. Additionally, the dataset was divided into training and validation sets. The resulting SEIRV model is depicted as Figure 3. The newly generated dataset has 16 columns (features) i.e., 11 parameters, 5 state variables and time. Rows where all cells contain zeros were also dropped. GridSearchCV is a handy utility from Python's sklearn module for hyperparameter tweaking ML models. It examines extensively over a defined variable grid and analyses the performance of the model through cross-validation to determine the optimal combination of hyperparameters. GridSearchCV was used for the following ML models: HGB, ENR, KNN, Ridge and Lasso regression.

Fig. 3: Second SEIRV Model

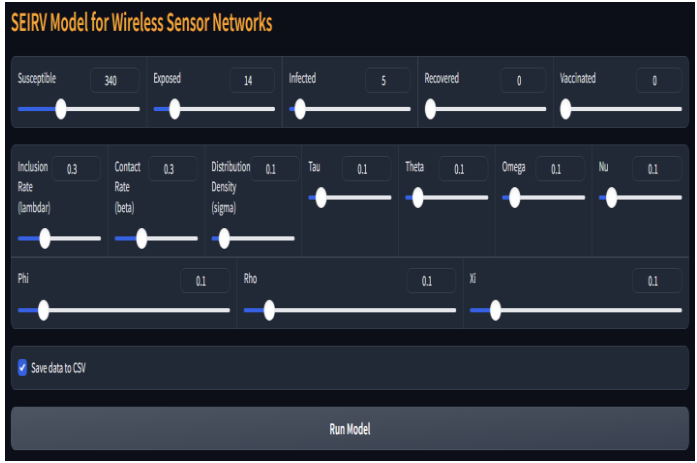

## IV. RESULTS AND DISCUSSIONS

In this section, the results of several experiments are presented. The error values for predicting the infected and recovered compartments using several algorithms were presented. Note that these were done for datasets generated using the two different approaches to synthetic dataset generation. The performance of the models were evaluated using the error metrics whice mathematical expressions are listed below.

### A. Results of the SEIRV NetLogo Model

Predictions of the infected and recovered compartments were conducted here. The table columns include: Algorithm: is the regression method under evaluationincluding HGB, RF, and XGBoost. Training time (TT) is the time (in seconds) required to train the ML model on the training dataset – a significant measure, particularly when comparing the efficiency of various algorithms. R-squared (train) is a coefficient of determination that reflects the degree to which the variables that are independent explain the variation of the variable that is dependent in the training set. Values near one indicate a better fit. In this scenario, XGBoost has the greatest R-squared score of 0.997, suggesting that it accounts for 99.7% of the variation in the training dataset. MAE (Train) is the average absolute difference between actual and predicted values in a training set. Lower numbers imply improved model performance. The RF model exhibits the smallest MAE of 27.057, indicating that it possesses the fewest average errors in predictions. MSE (train) is the average of the squares of the errors, which measures the mean squared disparity between actual and predicted values. Similarly, to MAE, lower numbers are preferable. The XGBoost model has the lowest MSE of 2771.659, suggesting the fewest average squared error. MAPE (train) (mean absolute percentage error) represents the average absolute percentage difference between anticipated and actual values. This statistic helps you grasp the inaccuracy in relative terms. The HG model had the smallest MAPE of 8.705%, implying that its predictions are around 8.7% off the mark.

For predictions of infected nodes using the train set, the following conclusions can be drawn from the table 1. XGBoost has the greatest R-squared value (0.997) with the smallest MSE (2771.659), indicating it is the best-performing model for most accurately predicting training data. HG additionally performed well, with an R-squared value of 0.993 and the smallest MAPE (8.705%), showing a reasonable balance between the training time and prediction accuracy. XGBoost has a low training time (0.157 seconds), which makes it more efficient than other models such as HGB (27.791 seconds) and MLP (9.369 seconds). Contrarily, the worst performers include SVR the highest errors and lowest Rsquared value. Followed by MLP; while not as poor as the SVR, it still has higher error values compared to other models and takes significant time to train. Tables 1, 2, 3 and 4.

TABLE I. INFECTED (TRAIN SET & TT) AT X-COR AND Y-COR =12

| Algorithm | TT | $R^2$ (Train) | MAE (Train) | MSE (Train) | MAPE (Train) |
|---|---|---|---|---|---|
| HGB | 27.791 | 0.993 | 50.201 | 8411.207 | 8.705 |
| RF | 3.372 | 0.995 | 27.057 | 6540.920 | 20.289 |
| XGBoost | 0.157 | 0.997 | 31.028 | 2771.659 | 10.007 |
| MLP | 9.369 | 0.953 | 153.052 | 60956.443 | 43.537 |
| DT | 0.020 | 0.996 | 2.685 | 4357.097 | 14.595 |
| SVR | 1.476 | 0.832 | 352.750 | 221296.662 | 100.200 |
| LiR | 0.023 | 0.944 | 168.385 | 73427.462 | 57.641 |
| LaR | 0.211 | 0.944 | 168.341 | 73427.999 | 57.641 |
| RiR | 0.107 | 0.944 | 168.351 | 73428.466 | 57.642 |
| KNN | 0.225 | 0.992 | 44.931 | 9889.519 | 19.056 |
| ENR | 0.314 | 0.944 | 168.322 | 73431.212 | 57.644 |

Note that the best number of neighbors is 3 for KNN. Additionally, the best alpha values in Tables 1 and 2 are 0.1, 1.00, 0.001. Best alpha values for in Tables 3 are 0.01, 1.00, 0.001; while in Table 4 they are 0.01, 0.1 and 0.001. In Table 5 they are 0.001, 0.1 and 0.001, where as in Table 6 they are 0.10, 0.10 and 0.001. In Tables 7 and 8 they are 0.01, 0.01 and 0.001 for LaR, RiR, and ENR, respectively.

For predictions of infected nodes, the following conclusions can be drawn from the table 2. The best Performing Models are XGB and KNN. In the validation set, XGBoost outperformed all other models when predicting infected nodes. It possesses a high R-squared value, which means that it describes 99.1% of the variation in the target variable. The relatively low MAE, MSE, and MAPE values indicate that its projections are very near to the true values, consequently rendering it extremely dependable for this job. KNN behind XGBoost in performance. It has a high R-squared value and somewhat superior MAE and MSE, indicating that it is likewise quite successful in predicting infected nodes. The somewhat greater MAPE compared to XGBoost indicates that, while KNN works well, its percentage error may vary slightly more. The worst-performing models are SVR and MLP for predicting infected nodes based on the validation set. The SVR has a relatively low R-squared value, implying that it explains just 83% of the variation. The large MAE, MSE, and MAPE values indicate that its predictions are significantly different from the actual values, showing a poor fit to the data and rendering it untrustworthy for this task. The MLP also performs badly compared to the other models. Although it possesses a decent R-squared value, both MAE and MSE are significantly greater compared to the best-performing models, suggesting that forecasts may be less accurate. The extremely high MAPE indicates great fluctuation in prediction accuracy, rendering it unsuitable for real-world use.

TABLE 2. INFECTED (VAL SET) AT XCOR/Y-COR =12

| Algorithm | R²(Val) | MAE(Val) | MSE (Val) | MAPE(Val) |
|---|---|---|---|---|
| HGB | 0.987 | 69.446 | 15817.441 | 5.935 |
| RF | 0.990 | 61.736 | 12611.169 | 9.236 |
| XGBoost | 0.991 | 59.864 | 10702.164 | 5.220 |
| MLP | 0.956 | 153.255 | 56592.832 | 46.756 |
| DT | 0.988 | 70.011 | 15076.086 | 6.065 |
| SVR | 0.830 | 362.024 | 219653.163 | 115.280 |
| LiR | 0.943 | 171.687 | 72630.021 | 76.749 |
| LaR | 0.943 | 171.640 | 72609.055 | 76.743 |
| RiR | 0.943 | 171.649 | 72604.683 | 76.741 |
| KNN | 0.992 | 58.573 | 10151.829 | 5.504 |
| ENR | 0.943 | 171.625 | 72582.489 | 76.736 |

TABLE 3. RECOVERED (TRAIN SET & TT) AT X-COR/Y-COR =12

| Algorithm | TT | R2 | MAE | MSE |
|---|---|---|---|---|
| HGB | 25.855 | 0.999 | 9.475 | 181.630 |
| XGBoost | 0.787 | 0.999 | 9.475 | |
| RF | 1.297 | 0.999 | 7.876 | 270.406 |
| MLP | 22.031 | 0.996 | 32.787 | 3438.825 |
| DT | 0.019 | 1.000 | 0.003 | 0.012 |
| SVR | 0.837 | 0.895 | 217.609 | 93097.517 |
| LiR | 0.038 | 0.988 | 53.272 | 10021.837 |
| LaR | 0.209 | 0.988 | 53.281 | 10021.840 |
| RiR | 0.114 | 0.988 | 53.288 | 10021.845 |
| KNN | 0.243 | 0.998 | 15.563 | 927.676 |
| ENR | 0.156 | 0.988 | 53.585 | 10024.821 |

TABLE 4. RECOVERED (VAL SET) AT X-COR/Y-COR =12

| Algorithm | R² (Val) | MAE (Val) | MSE (Val) |
|---|---|---|---|
| HGB | 0.997 | 25.365 | 2161.110 |
| XGBoost | 0.997 | 22.291 | 1862.547 |
| RF | 0.997 | 22.428 | 2296.891 |
| MLP | 0.996 | 33.276 | 3344.789 |
| DT | 0.996 | 26.145 | 3044.073 |
| SVR | 0.895 | 224.925 | 93812.878 |
| LiR | 0.987 | 56.385 | 10857.107 |
| LaR | 0.987 | 56.396 | 10856.828 |
| RiR | 0.987 | 56.403 | 10856.262 |
| KNN | 0.998 | 22.629 | 1591.428 |
| ENR | 0.987 | 56.738 | 10843.381 |

*B. Evaluating the Effect of Grid-based Coordinates*

The concepts of max-pxcor, min-pxcor, max-pycor, and min-pycor are frequently utilized in the context of grid-based simulations or spatial modelling, notably in NetLogo. These settings are helpful for specifying the grid limits within which agents (such as turtles and patches) may function. Understanding these coordinates aids the management of agent motions, determining visibility, and constraining interactions inside the given space. These parameters were changed to to -35, 35, -35 and 35, respectively, for our experiments. We experimented with variations in the coordinates to evaluate if it had any notable effect. These coordinates were motivated by the agent-based model in [18]. Tables 5, 6, 7 and 8 contains the error metrics for the change in grid-based coordinates (i.e., 35) using the first SEIRV model.

TABLE 5. INFECTED (TRAIN SET & TT) AT X-COR/Y-COR = 35

| Algorithm | TT | R² | MAE | MSE | MAPE |
|---|---|---|---|---|---|
| HGB | 27.195 | 0.983 | 39.165 | 4268.641 | 9.631 |
| RF | 2.150 | 0.995 | 17.352 | 1081.495 | 4.508 |
| XGBoost | 0.111 | 0.990 | 27.093 | 2337.676 | 9.317 |
| MLP | 16.042 | 0.935 | 68.235 | 16338.388 | 19.68% |
| DT | 0.033 | 0.999 | 0.412 | 25.845 | 0.127 |
| SVR | 2.0919 | 0.918 | 88.427 | 20600.475 | 25.209 |
| LiR | 0.025 | 0.889 | 111.723 | 28083.517 | 37.824 |
| LaR | 0.585 | 0.889 | 111.812 | 28085.506 | 37.740 |
| RiR | 0.112 | 0.889 | 111.755 | 28083.828 | 37.788 |
| KNN | 0.328 | 0.980 | 34.431 | 4902.065 | 8.734 |
| ENR | 0.281 | 0.888 | 113.098 | 28291.557 | 37.180 |

On the one hand, DT, HGB, RF, and XGBoost exhibited outstanding training performance measures, including high R-squared values as well as low MAE and MSE. SVR and MLP are less efficient, having lower R-squared values and larger error metrics for the prediction Recovered nodes on the train set. In contrast, HGB, XGBoost, and RF displayed great validation performance, as seen by high R-squared values and low error metrics (MAE and MSE). KNN also has the largest R-squared, showing superior predictive capability. Whereas, SVR is obviously the weakest model, with a low R-squared and a large error rate, suggesting that it does not generalize effectively to the validation set. While not the worst, the MLP performed poorly in comparison to the best algorithms for the prediction Recovered nodes on the validation set.

TABLE 6. INFECTED (VALIDATION SET) AT X-COR/Y-COR = 35)

| Algorithm | R² (Val) | MAE (Val) | MSE (Val) | MAPE (Val) |
|---|---|---|---|---|
| HGB | 0.968 | 49.597 | 8252.921 | 9.317% |
| RF | 0.971 | 46.591 | 7628.112 | 10.208% |
| XGBoost | 0.967 | 47.030 | 8634.036 | 14.25% |
| MLP | 0.946 | 66.499 | 14187.715 | 24.105% |
| DT | 0.941 | 59.8238 | 15664.175 | 9.445% |
| SVR | 0.933 | 85.762 | 17716.171 | 27.113% |
| LiR | 0.907 | 109.308 | 4692.761 | 38.995% |
| LaR | 0.907 | 109.356 | 24681.263 | 38.887% |
| RiR | 0.907 | 109.324 | 24687.556 | 38.950% |
| KNN | 0.963 | 50.157 | 9729.134 | 19.591% |
| ENR | 0.906 | 110.089 | 24765.697 | 38.066 |

For predicting the infected nodes using the train set and validation sets the results are as follows. DT and RF demonstrated in training with high R-squared scores and small error metrics, consequently being ideal for the regression task. SVR and MLP underperform owing to poor R-squared values and large error metrics. Additional refinement or other modelling approaches are recommended. DT and XGBoost have minimal training time and high performance, making them ideal for huge datasets. Linear, Lasso, Ridge, and ElasticNet regressions have greater error rates, indicating they may not be appropriate for this dataset. RF and HGB outperformed on the validation set, with high R-squared values as well as low error metrics, making them ideal for this prediction. Linear Regression, Lasso, Ridge, and ElasticNet regressions performed poorly, with high MAE and MAPE suggesting potential for improvement. KNN and XGBoost offer high performance and efficiency, while prediction accuracy may vary.

TABLE 7. RECOVERED (TRAIN SET & TT) AT X-COR/Y-COR = 35

| Algorithm | Training Time (Secs) | R² (Train) | MAE (Train) | MSE (Train) |
|---|---|---|---|---|
| HGB | 29.04 | 0.999 | 10.127 | 284.188 |
| XGBoost | 0.111 | 0.999 | 5.816 | 76.561 |
| RF | 2.142 | 0.999 | 4.250 | 54.903 |
| MLP | 14.906 | 0.998 | 13.145 | 576.055 |
| DT | 0.032 | 1.000 | 0.019 | 0.282 |
| SVR | 2.009 | 0.969 | 51.511 | 12620.915 |
| LiR | 0.001 | 0.997 | 23.551 | 1191.300 |
| LaR | 0.109 | 0.997 | 23.550 | 1191.300 |
| RiR | 0.105 | 0.997 | 23.551 | 1191.300 |
| KNN | 0.566 | 0.999 | 11.406 | 374.331 |
| ENR | 0.001 | 0.097 | 23.421 | 1191.998 |

DT is the best performer, with perfect prediction the capacity and low error alongside reduced training time. RF was also a strong performer, but requires more computation resources. ENR is the worst performer, with little explanatory strength and elevated error, demonstrating a need for a different modelling approach or better feature selection. SVR outperforms ElasticNet but falls short of the best models in terms of dependability, indicating that more tweaking or changes may be required. RF is the most accurate model for forecasting recovered instances due to its low error rates and near-perfect variability explanation. HGB provides comparable performance and is a solid option. SVR has greater error rates and worse dependability, whereas Linear Regression performs well but falls short of the top models' accuracy. These findings indicate a strong preference for ensemble approaches like as RF and HGB for superior prediction performance in this setting.

TABLE 8. RECOVERED (VAL SET) AT X-COR/Y-COR = 35

| Algorithm | R2 (Val) | MAE (Val) | MSE (Val) |
|---|---|---|---|
| HGB | 0.999 | 11.707 | 392.266 |
| XGBoost | 0.998 | 11.851 | 445.617 |
| RF | 0.999 | 11.054 | 320.803 |
| MLP | 0.998 | 13.394 | 594.749 |
| DT | 0.998 | 14.272 | 529.315 |
| SVR | 0.976 | 47.524 | 9864.253 |
| LiR | 0.997 | 23.100 | 1226.425 |
| LaR | 0.997 | 23.09 | 1226.408 |
| RiR | 0.997 | 23.099 | 1226.419 |
| KNN | 0.998 | 13.279 | 575.183 |
| ENR | 0.997 | 22.952 | 1224.93 |

*C. Results of the SEIRV Python Model*

Here, different algorithms were applied to the dataset generated by running the Python SEIRV model to understand how they perform while predicting the infected, recovered, and vaccinated compartments. Through inspection of the dataset, we noticed some zeros, which were dropped to eliminate issues during evaluation. Initially, the dataset used here has 39239 records, thus reducing the number of records to 24917 and 6 columns. The profile report showed the presence of skewness, and we first applied Box-Cox transformation, which failed. Therefore, we applied the Yeo-Johnson transformation (YJT) - a well suited method to zeros, negative and positive skewness, and it worked. StandardScaler and PowerTransformer was imported from sklearn.preprocessing library; the latter made it possible for us to use YJT. Furthermore, YJT was applied with and without GridSearchCV, which optimized parameter tuning for the regressors. NaN was obtained as R2 for the train and validation set when RF was utilized without GridSearchCV. Note that as was expected, the GridSearchCV tool added to the computation time. Additionally, outliers were removed using the zscore method to ease computational issues but it did not help as was expected.

## V. DISCUSSION

Note that the criteria for selecting best performing models include high $R^2$ and low MAE, MSE and MAPE, while the criteria for worst performing models include low $R^2$ and high MAE, MSE and MAPE, if they exist. It is necessary to define these criteria as it might be confusing considering that an algorithm such as HGB has a high training time of almost 25seconds, yet is among the best performers in the study. For model performance comparison, we use colored bars illustrating when the metrics are either high and low. Fig 3 shows the training times during the epidemic predictions. Figures 4, 5, 6, 7, 8, and 9 show the error metrics for the first dataset, while Figures 10 shows the error metrics for the second dataset. The predictions performed quite well, with low error metrics and high $R^2$ values (0.997, 1.000, 0.999, 1.000), indicating an effective fit to the training set. The validation values remained high (0.992, 0.998, 0.971, and 0.999), as is typical when evaluating model performance on unknown data. Training times indicate efficiency, particularly for simpler models like decision trees. During epidemic predictions using training data, XGBoost performed well with an R2 value of 0.997, equal to DT's results (1.000, 0.999, and 1.000). Furthermore, while predicting using the validation dataset, it returned values such as KNN (0.992 and 0.998) and RF (0.971 and 0.999). In contrast, SVR and ENR underperformed the other models. Specifically, EVR underperformed while predicting infected nodes using the training dataset. While SVR underperformed during the prediction of the infected and recovered nodes.

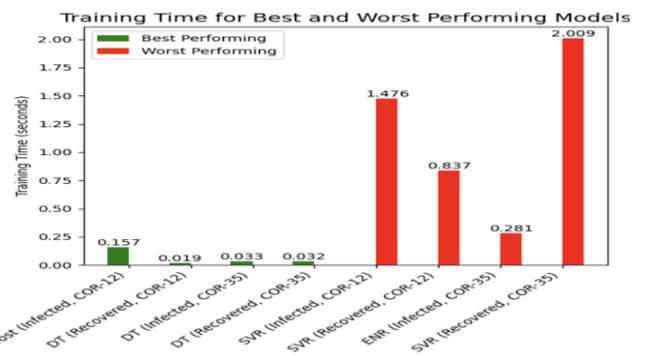

Fig 4. Training time during the predictions (Best Performing (BP))

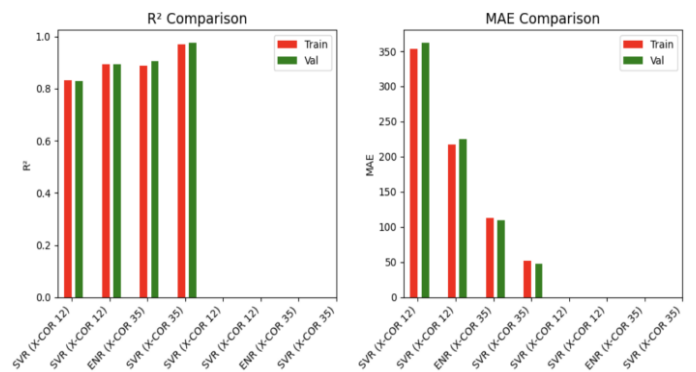

Fig 5. R Squared and MAE for training and validation set (BP)

On the second dataset derived using the second SEIRV model, our discussions were basically centered around $R^2$ values (Table 9 and 10). We found that the values of R-squared for the training and validation datasets in the regression analysis were remarkably high, suggesting that our model explained a sizable percentage of the variance in the target variables. In particular, the training set's R-squared values of 0.99 and validation set's results of 0.998 indicate an excellent model fit. The error metric; MAE, MSE, and RMSE were excessively large. However, it was also observed the target variable's scale, which is inclined toward extreme values and large volatility, is the cause of this disparity.

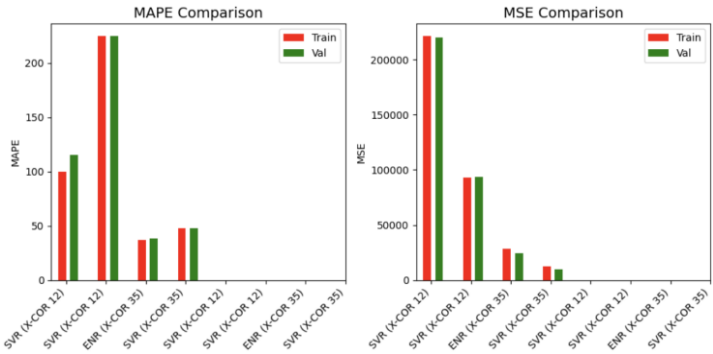
Fig 6. MAPE and MSE for training and validation set (BP)
Fig 7. R Squared and MAE for training and validation set (WP)

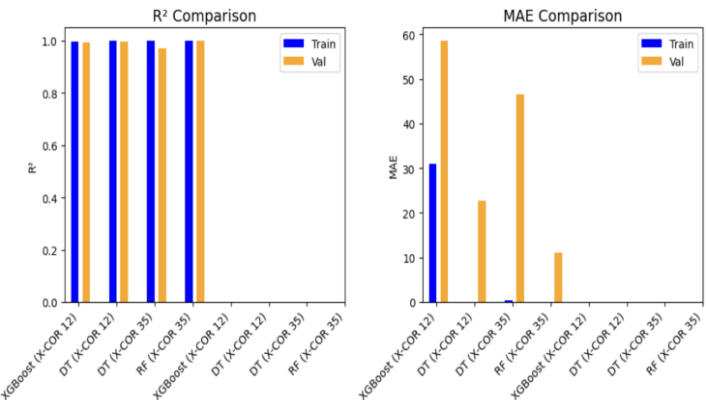
Fig 8. MAPE and MSE for training and val set (WP)

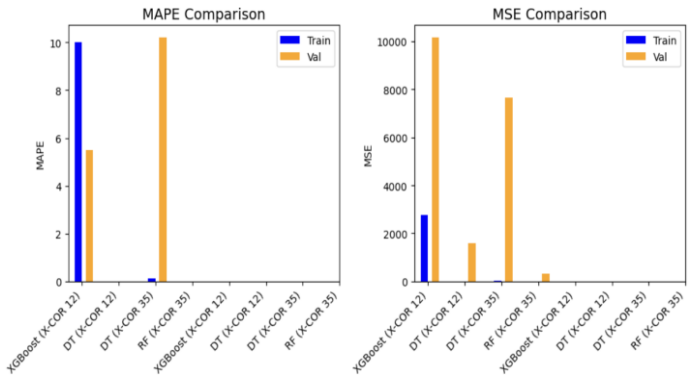

The $R^2$ values and training time were plotted as bar diagrams to graphically illustrate the dynamics of the second dataset obtained from the SEIRV model, wherein all the parameters were included in the experiments. Table 9 and Table 10 compares the efficiency and efficacy of each algorithm in predicting intended results, considering $R^2$ values and training times. From Fig 9, it is evident that the XGBoost and DT algorithms performed exceptionally well with perfect $R^2$ scores on the training set along with elevated values on the validation set. KNN showed moderate performance, particularly in validation, but MLP performs moderately. The LiR, LaR, RiR, and ENR, as well as HGB, have low $R^2$ values for validation and training, indicating they are unsuitable for this dataset. In addition, Fig 10 showed the top-performing algorithms; HGB, RF, and KNN, with high $R^2$ values on validation and training sets, suggesting their efficacy in predicting infected and recovered nodes. XGBoost and DT performed well too, but have significantly lower $R^2$ values than the top three. Contrastingly, the LaR, RiR, MLP, LiR, and ENR algorithms perform poorly on this dataset, with low $R^2$ scores for validation and training sets.

TABLE 9. INFECTED TRAINING & VALIDAITON SETS

| Algorithm | Training Time | $R^2$ (Train) | $R^2$ (Val) |
|---|---|---|---|
| **RF** | 132.24 | 0.99 | 0.99 |
| **XGB** | 123.74 | 1.00 | 0.99 |
| **DT** | 5.88 | 1.00 | 1.00 |
| **KNN** | 1.39 | 0.94 | 1.00 |
| **MLP** | 40.53 | 0.63 | 0.62 |
| **LiR** | 0.02 | 0.34 | 0.06 |
| **LaR** | 0.80 | 0.34 | 0.06 |
| **RiR** | 0.16 | 0.34 | 0.06 |
| **ENR** | 2.38 | 0.32 | 0.13 |
| **HGB** | 2.38 | 0.13 | 0.32 |

TABLE 10. RECOVERED TRAINING & VALIDAITON SETS

| Algorithm | Training Time | $R^2$ Train | $R^2$ Val |
|---|---|---|---|
| **RF** | 3.34 | 0.997 | 0.997 |
| **HGB** | 3.05 | 0.998 | 0.995 |
| **XGB** | 2.87 | 0.987 | 0.986 |
| **KNN** | 0.02 | 0.997 | 0.996 |
| **DT** | 0.11 | 0.996 | 0.995 |
| **MLP** | 0.39 | 0.39 | 0.41 |
| **LiR** | 0.01 | 0.11 | 0.10 |
| **LaR** | 0.01 | 0.11 | 0.10 |
| **RiR** | 0.01 | 0.11 | 0.10 |
| **ENR** | 0.01 | 0.10 | 0.09 |

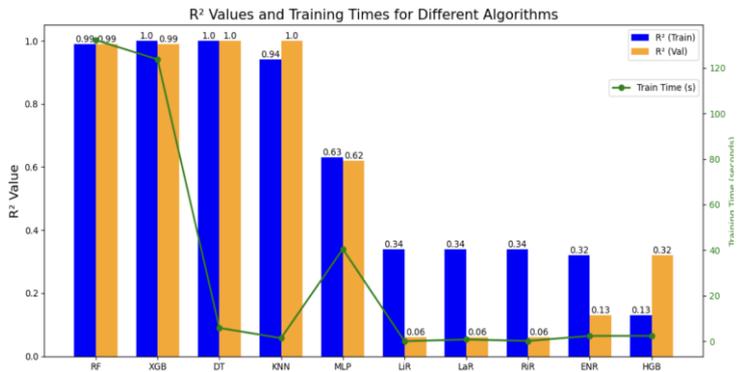
Fig 9. $R^2$ and training times for training and validation set (Infected)
Fig 10. $R^2$ and training times for training and validation set (Recovered)

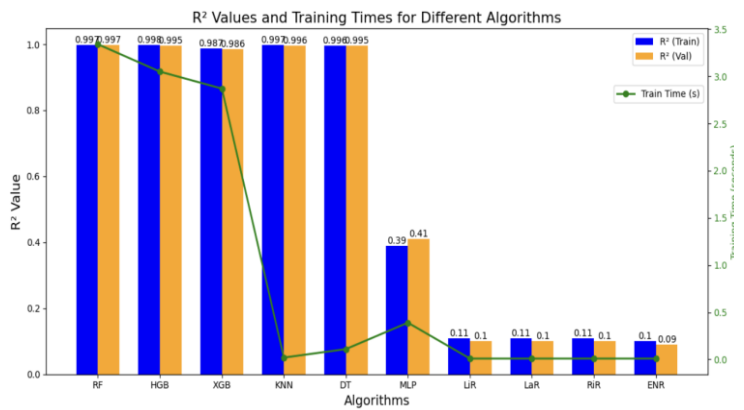

## VI. Conclusion and Future Direction

Within this work, an agent-based implementation of the susceptible-exposed-infected-recovered-vaccinated (SEIRV) mathematical model was developed and used for machine learning (ML) predictions. Two epidemic synthetic datasets were constructed and prepped for use with various ML algorithms using tools such as Netlogo's BehaviorSpace and Python. The conclusions derived from several experiments herein can assist in improved model selection and advances in predictive analytics for infection prediction, focusing on models that successfully balance prediction accuracy as well as training time. Summarily reiterating our experience here, the SEIRV model coded differently generated two datasets, one less complex and the other very complex. The less complex one had issues of persistently high correlations, while the very complex dataset had issues of skewness. These issues were addressed using several techniques. In the future, we would apply other types of transformations, such as logarithmic or square root, to the second dataset, since it seems more complex. Additionally, we would evaluate the impact of recurrent convolutional and recurrent neural networks.

## References

[1] X. Yu and Y. Xue, "Smart grids: A cyber–physical systems perspective," Proceedings of the IEEE, vol. 104, no. 5, pp. 1058–1070, 2016.

[2] S. Mendon¸ca, B. Dam´asio, L. C. de Freitas, L. Oliveira, M. Cichy, and A. Nicita, "The rise of 5g technologies and systems: A quantitative analysis of knowledge production," Telecommunications Policy, vol. 46, no. 4, p. 102 327, 2022.

[3] M. U. Saleem, M. R. Usman, M. A. Yaqub, A. Liotta, and A. Asim, "Smarter grid in the 5g era: Integrating the internet of things with a cyber-physical system," IEEE Access, 2024.

[4] C. Stamford. "Gartner survey shows rising concern of ai-enhanced malicious attacks as top emerging risk for enterprises for second consecutive quarter." (2024), [Online]. Available: https://www.gartner.com/en/newsroom/press(accessed: 25.09.2024).

[5] C. H. Nwokoye and V. Madhusudanan, "Epidemic models of maliciouscode propagation and control in wireless sensor networks: An indepth review," Wireless personal communications, vol. 125, no. 2, pp. 1827–1856, 2022.

[6] S. Chen, M. Hao, F. Ding, et al., "Exploring the global geography of cybercrime and its driving forces," Humanities and Social Sciences Communications, vol. 10, no. 1, pp. 1–10, 2023.

[7] M. Rhode, P. Burnap, and K. Jones, "Early-stage malware prediction using recurrent neural networks," computers & security, vol. 77, pp. 578–594, 2018.

[8] C. Nwokoye and I. Umeh, "Analytic-agent cyber dynamical systems analysis and design method for modeling spatio-temporal factors of malware propagation in wireless sensor networks," MethodsX, vol. 5, pp. 1373–1398, 2018.

[9] Y. Wu, C. Pu, G. Zhang, L. Li, Y. Xia, and C. Xia, "Epidemic spreading in wireless sensor networks with node sleep scheduling," Physica A: Statistical Mechanics and its Applications, vol. 629, p. 129 204, 2023.

[10] G. Liu, J. Li, Z. Liang, and Z. Peng, "Analysis of time-delay epidemic model in rechargeable wireless sensor networks," Mathematics, vol. 9, no. 9, 2021, issn: 2227-7390. doi: 10.3390/math9090978. [Online]. Available: https://www.mdpi.com/2227-7390/9/9/978.

[11] G. Liu, Z. Huang, X. Wu, Z. Liang, F. Hong, and X. Su, "Modelling and analysis of the epidemic model under pulse charging in wireless rechargeable sensor networks," Entropy, vol. 23, no. 8, p. 927, 2021.

[12] M. Srinivas, V. Madhusudanan, A. Murty, and B. Tapas Bapu, "A review article on wireless sensor networks in view of e-epidemic models," Wireless Personal Communications, vol. 120, pp. 95–111, 2021.

[13] I. Kotenko, "Agent-based modeling and simulation of cyber-warfare between malefactors and security agents in internet," in 19th European Simulation Multiconference "Simulation in wider Europe, 2005.

[14] I. Kotenko, "Chapter agent-based modeling and simulation of network infrastructure cyber-attacks, cooperative defense mechanisms," 2010.

[15] M. Niazi and A. Hussain, "Agent-based tools for modeling and simulation of self-organization in peer-to-peer, ad hoc, and other complex networks," IEEE Communications Magazine, vol. 47, no. 3, pp. 166–173, 2009.

[16] E. Mojahedi and M. A. Azgomi, "Modeling the propagation of topologyaware p2p worms considering temporal parameters," Peer-to-Peer Networking and Applications, vol. 8, pp. 171–180, 2015.

[17] K. Wasti, "Usability of multi-agent simulators in simulation of wireless networks," M.S. thesis, K. Wasti, 2014.

[18] M. A. Niazi and A. Hussain, "A novel agent-based simulation framework for sensing in complex adaptive environments," IEEE Sensors Journal, vol. 11, no. 2, pp. 404–412, 2010.

[19] N. Shone, T. N. Ngoc, V. D. Phai, and Q. Shi, "A deep learning approach to network intrusion detection," IEEE transactions on emerging topics in CI, vol. 2, no. 1, pp. 41–50, 2018.

[20] S. Chawla, Deep learning based intrusion detection system for Internet of Things. University of Washington, 2017.

[21] L. Xiaofeng, Z. Xiao, J. Fangshuo, Y. Shengwei, and S. Jing, "Assca: Api based sequence and statistics features combined malware detection architecture," Procedia Computer Science, vol. 129, pp. 248–256, 2018.

[22] A. Hijazi, A. El Safadi, and J.-M. Flaus, "A deep learning approach for intrusion detection system in industry network.," in BDCSIntell, 2018, pp. 55–62.

[23] J. Kang, S. Jang, S. Li, Y.-S. Jeong, and Y. Sung, "Long short-term memory-based malware classification method for information security," Computers & Electrical Engineering, vol. 77, pp. 366–375, 2019.

[24] X. Fang, M. Xu, S. Xu, and P. Zhao, "A deep learning framework for predicting cyber attacks rates," EURASIP Journal on Information security, vol. 2019, pp. 1–11, 2019.

[25] G. Thamilarasu and S. Chawla, "Towards deep-learning-driven intrusion detection for the internet of things," Sensors, vol. 19, no. 9, p. 1977, 2019.

[26] J. Ren, J. Guo, W. Qian, H. Yuan, X. Hao, and H. Jingjing, "Building an effective intrusion detection system by using hybrid data optimization based on machine learning algorithms," Security and communication networks, vol. 2019, no. 1, p. 7 130 868, 2019.

[27] A. Boukhalfa, A. Abdellaoui, N. Hmina, and H. Chaoui, "Lstm deep learning method for network intrusion detection system," International Journal of Electrical and Computer Engineering, vol. 10, no. 3, p. 3315, 2020.

[28] M. Almseidin, M. Alzubi, S. Kovacs, and M. Alkasassbeh, "Evaluation of machine learning algorithms for intrusion detection system," in 2017 IEEE 15th international symposium on intelligent systems and informatics (SISY), IEEE, 2017, pp. 000 277–000 282.

[29] L. Wuke, Y. Guangluan, and C. Xiaoxiao, "Application of deep extreme learning machine in network intrusion detection systems," IAENG Int. J. of Computer Science, vol. 47, no. 2, pp. 136–143, 2020.

[30] K. Kim, "Intrusion detection system using deep learning and its application to wi-fi network," IEICE TRANSACTIONS on Information and Systems, vol. 103, no. 7, pp. 1433–1447, 2020.

[31] Y. Wang, S. Wang, and G. Tong, "Learning the propagation of worms in wireless sensor networks," in International Wireless Internet Conference, Springer, 2022, pp. 102–115.

[32] E. S. Alomari, R. R. Nuiaa, Z. A. A. Alyasseri, et al., "Malware detection using deep learning and correlation-based feature selection," Symmetry, vol. 15, no. 1, p. 123, 2023.

[33] P. K. Bediako, Long short-term memory recurrent neural network for detecting ddos flooding attacks within tensorflow implementation framework. 2017.

[34] J. Kim, J. Kim, H. L. T. Thu, and H. Kim, "Long short term memory recurrent neural network classifier for intrusion detection," in 2016 international conference on platform technology and service (PlatCon), IEEE, 2016, pp. 1–5.

[35] R. C. Staudemeyer, "Applying long short-term memory recurrent neural networks to intrusion detection," South African Computer Journal vol. 56, no. 1, pp. 136–154, 2015